\pdfoutput=1
\documentclass[10pt,twocolumn,letterpaper]{article}

\usepackage[pagenumbers]{cvpr} 

\usepackage{graphicx}
\usepackage{amsmath}
\usepackage{amssymb}
\usepackage{booktabs}
\usepackage{placeins}
\usepackage{listings}
\usepackage{dirtree}

\usepackage[pagebackref,breaklinks,colorlinks]{hyperref}

\hypersetup{
    colorlinks = true,
    linkcolor = blue
}

\usepackage[capitalize]{cleveref}
\crefname{section}{Sec.}{Secs.}
\Crefname{section}{Section}{Sections}
\Crefname{table}{Table}{Tables}
\crefname{table}{Tab.}{Tabs.}

\def\cvprPaperID{*****} 
\def\confName{arXiv}
\def\confYear{2023}

\begin{document}

\title{VALERIE22 - A photorealistic, richly metadata annotated dataset of urban environments\\[1em]
  \includegraphics[width=0.6\textwidth]{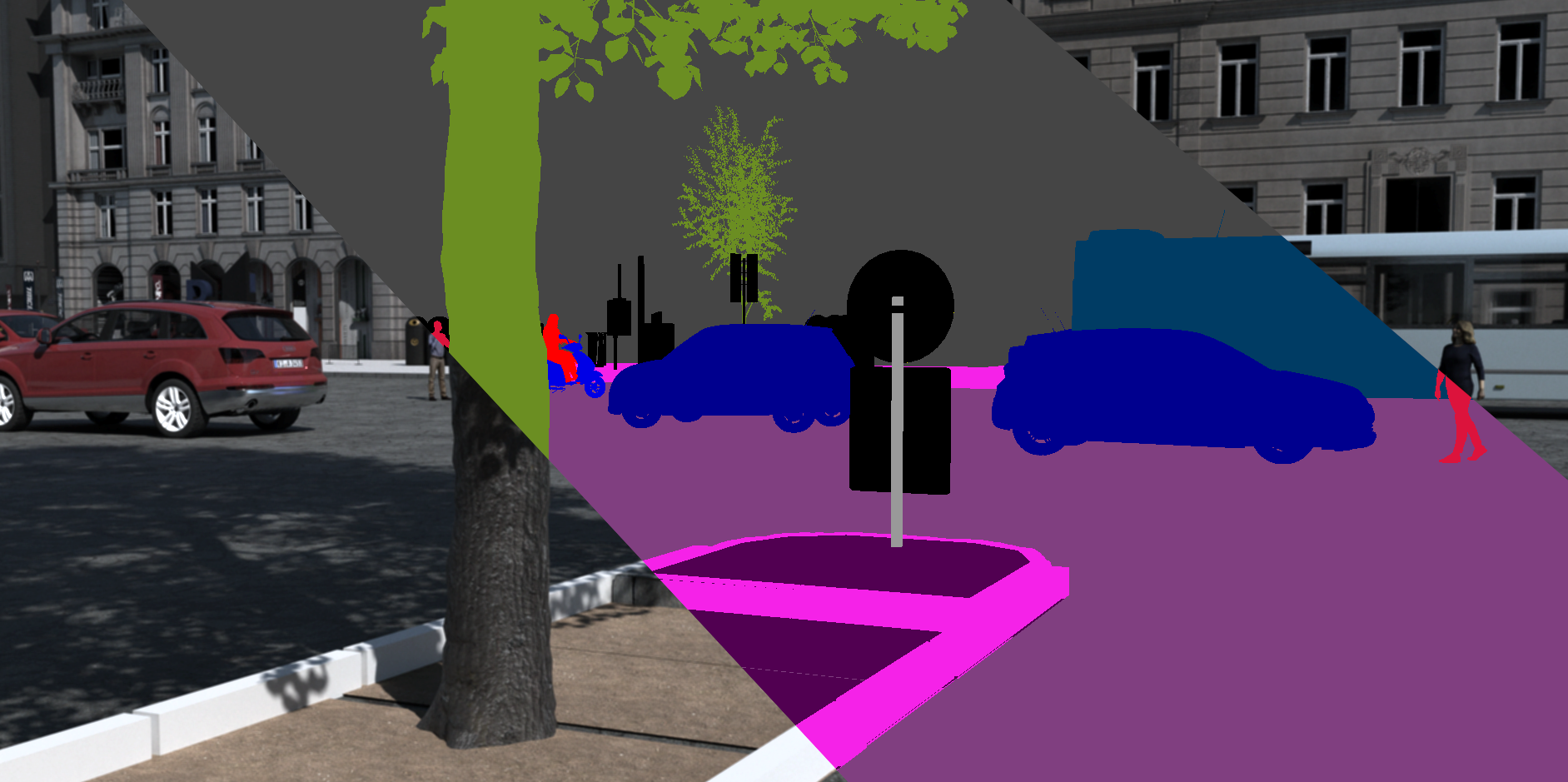}
  \includegraphics[width=0.25\textwidth]{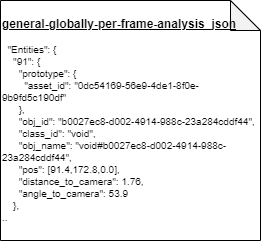}
}

\author{Oliver Grau\\
Intel Labs\\
{\tt\small oliver.grau@intel.com}
\and
Korbinian Hagn\\
Intel Labs\\
{\tt\small korbinian.hagn@intel.com}
}

\maketitle

\begin{abstract}
  The VALERIE tool pipeline is a synthetic data generator \cite{Grau2022} developed with the goal to contribute to the understanding of domain-specific factors that influence perception performance of DNNs (deep neural networks). This work was carried out under the German research project \emph{KI Absicherung} in order to develop a methodology for the validation of DNNs in the context of pedestrian detection in urban environments for automated driving.

  The \textit{VALERIE22} dataset was generated with the VALERIE procedural tools pipeline providing a photorealistic sensor simulation rendered from automatically synthesized scenes. The dataset provides a uniquely rich set of metadata, allowing extraction of specific scene and semantic features (like pixel-accurate occlusion rates, positions in the scene and distance + angle to the camera). This enables a multitude of possible tests on the data and we hope to stimulate research on understanding performance of DNNs.

  Based on performance metric a comparison with several other publicly available datasets is provided, demonstrating that \textit{VALERIE22} is one of best performing  synthetic datasets currently available in the open domain. \footnote{Available here: \url{https://huggingface.co/datasets/Intel/VALERIE22}}

\end{abstract}



\section{Introduction}
\label{sec:intro}

Recently, great progress has been made in applying machine learning techniques to deep neural networks to solve perceptional problems. Automated vehicles (AV) are a recent focus as an important application of perception from cameras and other sensors, such as LIDAR and Radar \cite{yurtsever2020survey}. Although the current main effort is on developing the hardware and software to implement the functionality of AVs, it will be equally important to demonstrate that this technology is safe.

The German collaborative research project \emph{KI Absicherung} \cite{kiaHOME} was a cross industry and academia effort to develop a methodology for the validation of DNNs in the context of pedestrian detection in urban environments for automated driving. Specifically, one important goal of that project was to make the safety aspects of ML-based perception functions predicable. As one important research stream of this project synthetic data generation was used as a base, as this allows full control over domain-specific scene parameters and the ability to generate parameter variations of these. Further, additional metadata annotations were specified and automated computation of these were added to the synthesis pipeline.

The VALERIE tools pipeline was developed as a research tool to improve quality of data synthesis and to get an understanding of factors that determine the domain gap between synthetic and real datasets. For that a powerful synthesis pipeline has been developed, which allows the fully automated creation of complex urban scenes. In this paper we only summarize some of the functionalities of the VALERIE synthesis pipeline and focus on a description of the (meta-)data formats of the \textit{VALERIE22} dataset that was generated with the tool chain. More details on the synthesis tools can be found in \cite{Grau2022}.

Additionally, we present evaluation results to assess the quality of our synthetic data compared to other synthetic datasets in the autonomous driving domain.

\subsection{Related work}
\label{sec:related}

In \cite{Grau2022} we suggest a computational data synthesis approach for deep validation of perception functions based on parameterized synthetic data generation. We introduce a multi-stage strategy to sample the input domain and to reduce the required vast amount of computational effort. This concept is an extension and generalization of our previous work on parameterization of the scene parameters of concrete scenarios. We extended this parameterization by a probabilistic scene generator to widen the coverage of scenario spaces and a more realistic sensor simulation. These approaches were used to generate the scenes and data in the \textit{VALERIE22} dataset.

Techniques to capture and render models of the real world have been matured significantly over the last decades. Computer generated imagery (CGI) is increasingly popular for training and validation of deep neural networks (DNNs) as synthetic data can avoid privacy issues found with recordings of members of the public and can automatically produce  ground truth data at higher quality and reliability than costly manually labeled data. Moreover, simulations allow synthesis of rare scene constellations helping validation of products targeting safety critical applications, specifically automated driving.
Because of the progress in visual and multi-sensor synthesis, now building systems for validation of these complex systems in the data center becomes not only feasible but also offers more possibilities for the integration of intelligent techniques in the engineering process of complex applications.

The use of synthesized data for development and validation is an accepted technique and has been also suggested for computer vision applications (e.g. \cite{Burger1995}).
Several methodologies for verification and validation of AVs  have been developed \cite{Kalra16,junietz2018evaluation,damm2018exploiting} and commercial options exist.\footnote{For example Carmaker from IPG  or PreScan from TASS International.} These tools were originally designed for virtual testing of automotive functions, like braking systems and then extended to provide simulation and management tools for virtual test drives in virtual environments.
They provide real-time capable models for vehicles, roads, drivers, and traffic which are then being used to generate test (sensor) data as well as APIs for users to integrate
the virtual simulation into their own validation systems.


Recently, specifically in the domain of driving scenarios, game engines have been adapted
\cite{wymann2000torcs,richter2016playing}. Another virtual simulator system, which gained popularity in the research community is  CARLA  \cite{DBLP:journals/corr/abs-1711-03938}, also based on a commercial game engine (Unreal4 \cite{Unreal4}).
Although game engines provide a good starting point to simulate environments, they usually only offer a closed rendering set-up with many trade-offs balancing between real-time constraints and a subjectively good visual appearance to human observers. Specifically, the lighting computation in this rendering pipelines is limited and does not produce physically correct imagery. Instead, game engines only deliver fixed rendering quality typically with 8bit per RGB color channel and only basic shadow computation.

In contrast, physical-based rendering techniques have been applied to the generation of data for training and validation, like in the \textit{Synscapes} dataset\cite{wrenninge2018synscapes}.
For our experimental work we use the physical-based open source Blender Cycles renderer\footnote{\url{https://www.blender.org/}} in high dynamic range (HDR) resolution. 

The effect of sensor and lens effects on perception performance has only been limited studied.
In \cite{Carlson_2018_ECCV_Workshops,8954694} the authors are modeling camera effects to improve synthetic data for the task of bounding box detection. Metrics and parameter estimation of the effects from real camera images are suggested by \cite{Liu2020NeuralNG} and \cite{carlson2019sensor}.
A sensor model including sensor noise, lens blur, and chromatic aberration was developed based on real data sets \cite{HagnK2021} and integrated into our validation framework.

Looking at virtual scene content, most recent simulation systems for validation of complete AD system include simulation and testing of the ego-motion of a virtual vehicle and its behavior. The used test content or scenarios are therefore aiming to simulate environments with a large extension and are virtually driving a high number of test miles (or km) in the virtual world provided \cite{Menzel18,wen2020scenario,damm2018exploiting}. This might be a good strategy to validate full AD stacks, one problem for validation of perception systems is the limited coverage of data testing critical and performance limiting factors.


A more suitable approach is to use probabilistic grammar systems \cite{devaranjan2020metasim2,wrenninge2018synscapes} to generate 3D scenarios which include a catalog of different object classes and places them relative to each other to cover the complexity of the input domain. The \textit{VALERIE22} dataset demonstrates the effectiveness of our  probabilistic grammar system together with our previous scene parameter variation \cite{SyedShar2020} with a novel multi-stage strategy.
This approach allows to systematically test conditions and relevant parameters for validation of the perceptional function under consideration in a structured way.

The remainder of this contribution is structured as the following: The next section will give an outline of our synthesis approach and a description of the generated meta-data.
In section \ref{sec:eval} we will give a comparison of \textit{VALERIE22} with a number of publicly available real and synthetic datasets.

\section{VALERIE data synthesis pipeline}
\label{sec:valerie}

\begin{figure}[ht!]
  \centering
  \includegraphics[width=0.95\linewidth]{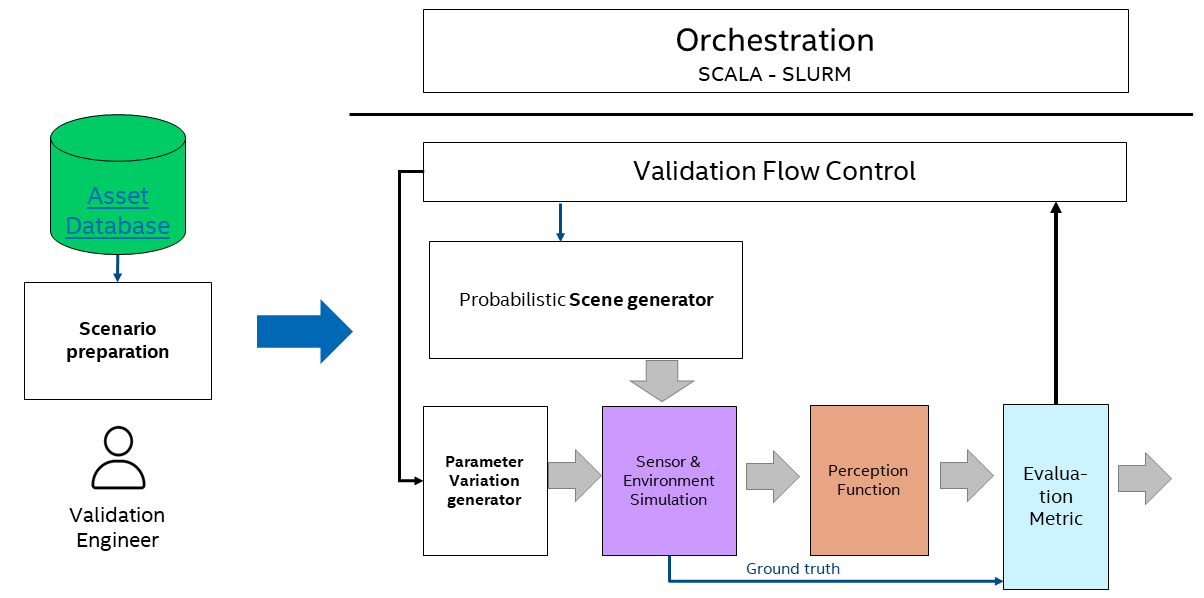}
  \caption{Overview of VALERIE pipeline flow. }
  \label{fig:overview}
\end{figure}

VALERIE is composed of several modules, as depicted in fig. \ref{fig:overview}. The validation flow control is in principle  designed to run automated validation strategies in a data center, with the help of the 'SCALA' orchestration module based on slurm\footnote{\url{https://slurm.schedmd.com/documentation.html}}. A description of the concept of these modules is outside the scope of this paper, see \cite{Grau2022} for more details. The aim in here is to only give an overview over some of the modules in the data synthesis part, so that the reader is able to understand  the features of the dataset and how to identify objects in the rendered frames.

\subsection{Computation of synthetic data}
\label{sec:syntheticdata}

Synthetic data is generated with graphics methods. Specifically for color (RGB) images, there are many software systems available, both commercially and as open source. For the generation of the dataset described in this paper Blender was used as a base to import, edit, and rendering of 3D content.

\begin{figure}[ht!]
  \centering
  \includegraphics[width=0.95\linewidth]{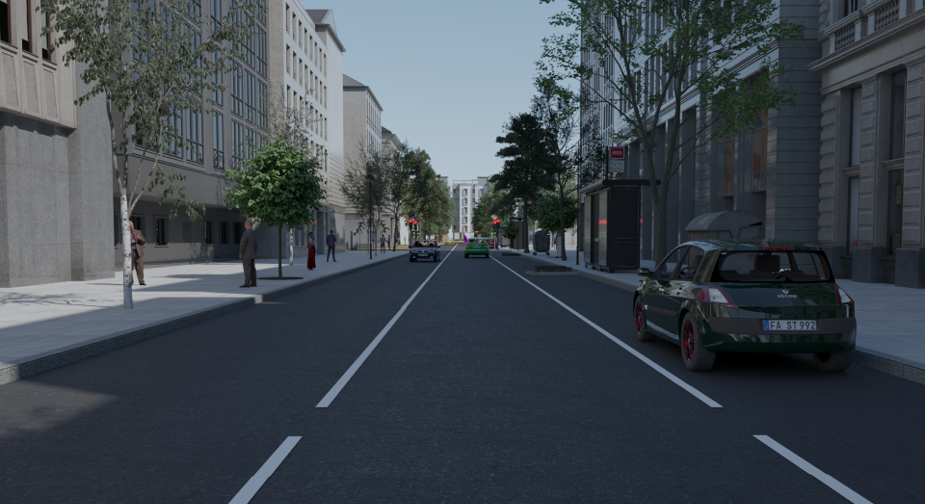}
  \includegraphics[width=0.95\linewidth]{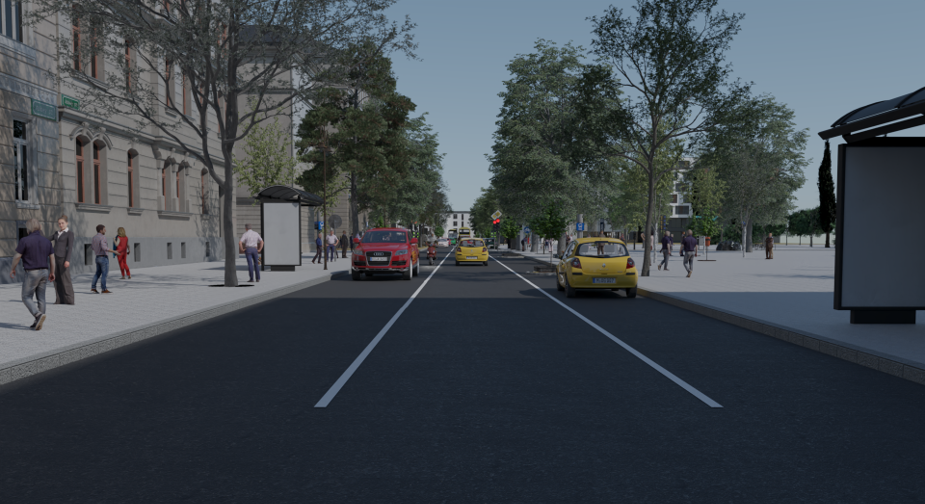}
  \includegraphics[width=0.95\linewidth]{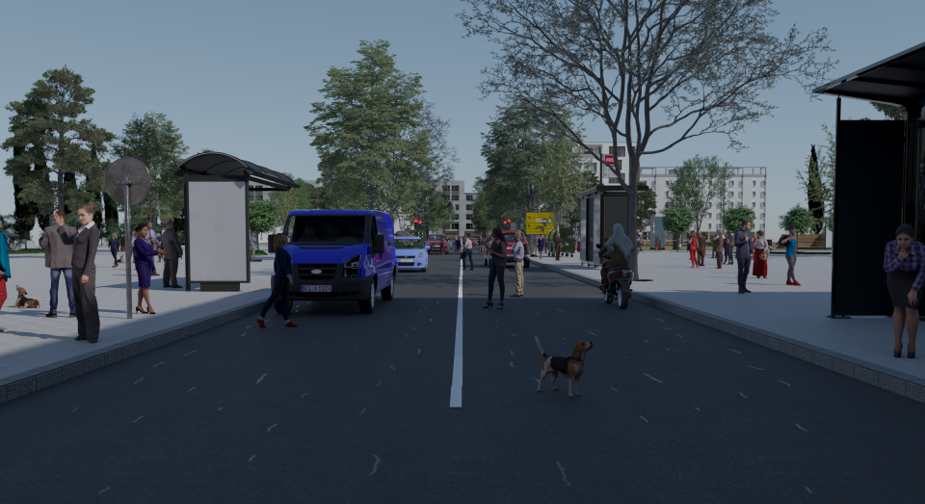}

  \caption{Variation of density of pedestrians in the street and on side walk (top) low, to high (bottom). }
  \label{fig:density}
\end{figure}

The generation of highly varied synthetic data involves the following steps:

\begin{enumerate}
  \item A 3D scene model with a city model is generated using a terrain/street generator. Parameters like width of a street and pavement, type of segment (e.g. tall houses, sub-urban residential, green/park, place, etc.) and materials for roads, sidewalks, segments are generated based on a scene description. Alongside this process the semantic information about the types and geometry of the segments is passed as input to the next step.
  \item A placement step is inserting 3D assets, like cars, vegetation, road elements and pedestrians into the scene. This placement is inserting objects based on a density declaration (per segment) and a list of assets for this type of segment (e.g. road, sidewalk, etc.). The result is a complete scene. Fig. \ref{fig:density} shows examples of scenes with a variation of person densities.

  \item (optionally) a set of scene parameters can be varied before each rendering pass. This includes position of objects, cameras and time-of-the-day (to vary the sun position) and many more.

\end{enumerate}

The dataset contains a multitude of additional metadata. For example all objects in the scenes are tagged with an identifier (see next section) and semantic and scene information, like position in the scene and distance + angle to the camera is documented in form of json files. This enables a multitude of possibilities to analyze the data and we hope to stimulate research on understanding performance of DNNs with our dataset.

\subsection{Assets and object instances}
\label{sec:assets}

The assets\footnote{An asset here means a 3D model or 2D texture.} in the asset database (left side in fig. \ref{fig:overview}) have a unique identifier in form of a UUID (Universally Unique IDentifier). This identifier is used in the scene description either explicitly (for static objects) or in selection lists used by the probabilistic scene generator.

The asset id\footnote{id == identifier for brevity.} is also used to identify objects in the rendered frames.  The dataset contains metadata files (json format) with a list of objects and their asset ids. Objects are also identified with a specific UUID. This is depicted in fig. \ref{fig:instanceids}. In the appendix, section on \emph{Metadata} 
an example json file is listed. The "entities" key, in this example "91" is an integer and corresponds to the instance label (see below) of the instance ground truth. With the help of the scene metadata files and the unique UUIDs of the assets it is possible to identify assets in the rendered scene. This can be used for statistical purposes or to retrieve more information from the asset database (not included in the dataset).

\begin{figure}[ht!]
  \centering
  \includegraphics[width=0.95\linewidth]{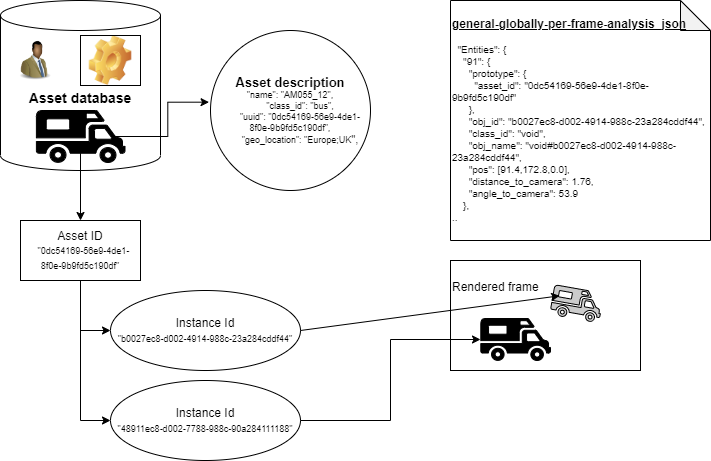}
  \caption{Object identifiers allow tracking of object instances through the rendered frames and metadata. }
  \label{fig:instanceids}
\end{figure}

The scene composition and also the used assets in \textit{VALERIE22} are European, e.g. the traffic signs and road markings are German. The types of houses are also mainly European style.



\subsection{Ground truth and metadata}
\label{sec:metadata}

The \textit{VALERIE22} dataset provides a very rich set of metadata annotations and ground truth:

\begin{itemize}
  \item pixel-aligned class groups (semantic label image)
  \item pixel-aligned object instances (label image)
  \item object 2D bounding box
  \item object 3D bounding box
  \item object position and orientation, angle and distance to camera
  \item object occlusion (only for person class)
  \item scene parameters, specifically time-of-the-day and sun (illumination)
  \item camera parameter, including pose in scene
\end{itemize}

The labels for object classes will be mapped to a convention used in annotation formats and follows the Cityscapes convention \cite{cordts2016cityscapes} for training and evaluation of the perception function.  The 2D image of a scene is computed along with the ground truth extracted from the modeling software rendering engine.

\subsection{Sensor Simulation}
\label{sec:sensor_simulation}

We implemented a sensor  model to simulate real sensor behavior.
The module works on HDR images in linear RGB space and floating point resolution as provided by the Blender Cycles renderer.

\begin{figure}[ht!]
  \centering
  \includegraphics[width=0.95\linewidth]{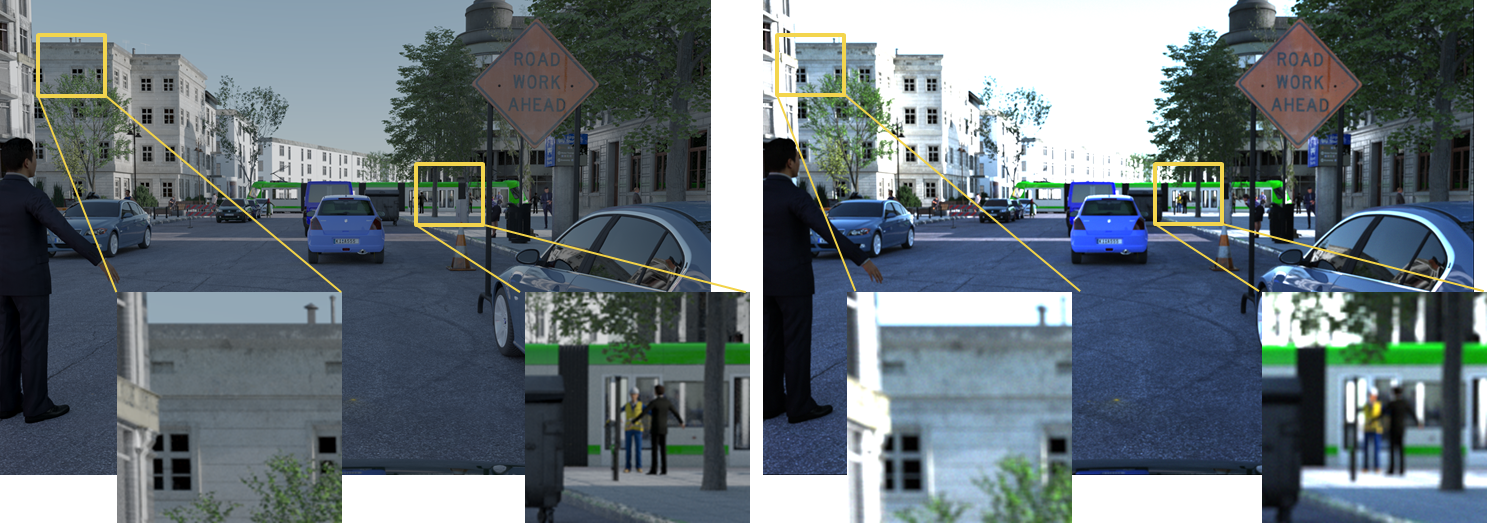}
  \caption{Realistic sensor effect simulation, (left) standard Blender tone-mapped output, (right) the sensor simulation output. }
  \label{fig:sensorsimulation}
\end{figure}

We simulate a camera error model by applying \textbf{sensor noise}, as added Gaussian Noise (mean=0, variance: free parameter) and a automatic, histogram-based exposure control (linear tone-mapping), followed by non-linear \textbf{Gamma correction}. Further, we
simulate the following lens artifacts \textbf{chromatic aberration}, and \textbf{blur}. Fig. \ref{fig:sensorsimulation} shows a comparison of the standard tone-mapped 8bit RGB output of Blender (left) with our sensor simulation (right). The parameters were adapted to approximate the camera characteristic of Cityscape images. The images do not only look more realistic for the human eye, they also are further closing the domain gap between the synthetic and real data  (for details see \cite{HagnK2021}).

\subsubsection{Sampling of variable parameter}
\label{subsec:validation-parameter}

Variations in the dataset were created by linear stepping through a parameter interval or random sampling of these. Examples are time-of-the-day to control the sun settings or position and orientation of the camera. The parameters used in variation runs are documented in a json file with the actual parameter variations. However, the sun camera parameters are also documented in the 'per-frame-analysis' file.

\section{Evaluation}
\label{sec:eval}

To evaluate the quality of our dataset  we conducted several experiments using the semantic segmentation task. We compare the segmentation performance of a \texttt{DeeplabV3+} model trained on our synthetic data and compare the performance with models trained on several synthetic datasets. The performance of these models is then evaluated on five different real world automotive segmentation datasets. Use cases of our metadata include improved training and identification of impairing factors (for more details see \cite{HagnK2022ECCV,HagnK2022}).

Next, we investigated the segmentation performance on the person class of the \textit{CityPersons} dataset if we train the model on subsets of our dataset. We additionally evaluated the person class performance with models trained on subsets of the \textit{SynPeDS} dataset \cite{StaunerT2022} provided by the KI Absicherung project\footnote{Currently a publication of the SynPeDS dataset is under preparation, see \url{https://www.ki-absicherung-projekt.de/}}.
Finally, we investigated how the performance of the models differs for the number of unique person assets used to create the datasets and their subsets.

Lastly, we investigated how the number of training images influences the segmentation performance. Again we trained on subsets of our dataset and the \textit{SynPeDS} dataset and evaluated the segmentation performance on all classes with the \textit{DeeplabV3+} segmentation model.

\subsection{Computation and evaluation of perceptional functions}
\label{sec:computeperc}

State-of-the-art perception functions consists of a multitude of different approaches considering the wide range of different tasks. For experiments presented in this chapter, we are considering the task of semantic segmentation. In this task, the perception function segments an input image into different objects by assigning a semantic label to each of the input image pixels. One of the main advantages of semantic segmentation is the visual representation of the task which can be easily understood and analyzed for flaws by a human.

In this work, we considered the \texttt{DeeplabV3+} model which originated from \cite{chen2017deeplab} and utilizes a \texttt{ResNet101} backbone.

We compare our dataset to three different synthetic datasets. The first dataset
is the synthetic dataset \textit{SynPeDS} \cite{StaunerT2022} consisting of urban street scenes inspired by the preceding two real-world datasets. The second dataset is the \textit{GTAV} dataset \cite{richter2016playing}, created by sampling data from the 3D game of the same name. Last, the \textit{Synscapes} dataset \cite{wrenninge2018synscapes} which is intended to synthetically re-create charateristics of the \textit{Cityscapes} dataset is considered.

To compare our dataset we train segmentation models on each of these datasets and evaluate the segmentation performance on five real-world datasets.
The first dataset is the \textit{Cityscapes} dataset \cite{cordts2016cityscapes}, a collection of European urban street scenes in the daytime with good to medium weather conditions. The second dataset
is the A2D2 by \cite{aev2019}, similar to the \textit{Cityscapes} dataset it is a collection of German  urban street scenes and additionally it has sequences from driving on a freeway. The third dataset is the \textit{BDD100K} dataset \cite{Yu2019} a diverse dataset recorded in North-America at diverse weather conditions. Next, the \textit{India Driving Dataset} dataset \cite{varma2018idd}, which was recorded in India and contains entirely different street scenes compared to the European or American datasets. Last, the \textit{Mapillary Vistas} dataset \cite{MVD2017}, a world wide dataset with emphasis on northern America.
All of these datasets are labeled on a subset of 11 classes which are alike in these datasets to provide comparability between the results of the different trained and evaluated models.

To measure the performance of the task of semantic segmentation the mean Intersection over Union (mIoU) from the COCO semantic segmentation benchmark task is used \cite{shelhamer2016fully}.
The mIoU is denoted as the intersections between predicted
semantic label classes and their corresponding ground truth divided by the union of the same, averaged over all classes.

We showed in our previous work  how to use the extensive metadata accompanied to our dataset to detect data biases in person detectors due to the underlying training data used to train the bounding box detectors \cite{HagnK2022ECCV}.

Another work investigated the usage of the metadata to calculate visual impairing factors, i.e., factors that lead to detrimental detection performance of a person detector such as increased occlusion or decreased contrast. Re-training a person detector with a focus on harder to detect samples, according to these factors, improves the overall detection performance \cite{HagnK2022}.

\subsubsection{Cross domain evaluation}

To demonstrate the quality of our synthetic dataset we conducted several cross-domain performance experiments with other real-world automotive and synthetic datasets. This cross-domain performance
analysis is also commonly referred to as generalization distance. We trained a DeeplabV3+ model on our \textit{VALERIE22} dataset, as well as for the \textit{SynPeDS}, the \textit{GTAV} and the \textit{Synscapes} dataset. Next, we evaluated the segmentation performance on real-world datasets \textit{A2D2}, \textit{BDD100K}, \textit{Cityscapes}, \textit{IDD} and \textit{Mapillary Vistas}.

As the real-world and synthetic datasets do not have exactly the same semantic annotation format,
the segmentation models were trained on a subset of 11 labels per dataset to ensure consistency of
classes across. The labels are defined as follows: Road and sidewalk incorporate the road-markings and the curb respectively. Further,
the building, sky, car and truck classes are used, which are consistent across
these datasets. Pole, traffic light and traffic sign classes are mapped
from similar sub-classes in the used datasets, e.g., utility pole in \textit{Mapillary Vistas}.
The vegetation class consists of the \textit{Cityscapes} sub-classes terrain, i.e., plants
covering the ground, and the original vegetation class, i.e., trees
and bushes. Last, the person class is defined as all humans in the
dataset, e.g., pedestrians and riders.

\begin{figure}[!ht]
  \centering
  \includegraphics[width=1.0\columnwidth]{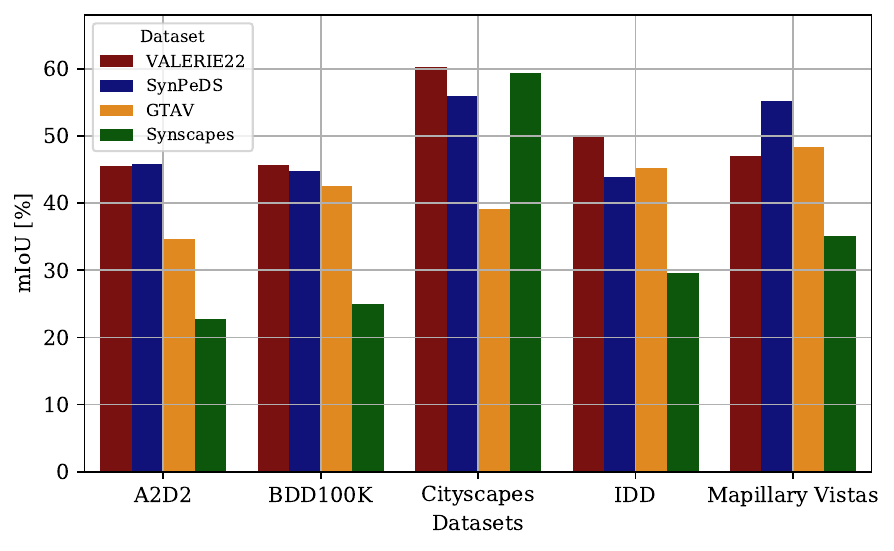}
  \caption{Cross-domain segmentation performance of synthetic datasets \textit{VALERIE22}, \textit{SynPeDS}, \textit{GTAV} and \textit{Synscapes} evaluated on real world datasets \textit{A2D2}, \textit{BDD100K}, \textit{Cityscapes}, \textit{IDD} and \textit{Mapillary Vistas}.}
  \label{fig:cross_domain}
\end{figure}

The mIoU cross-domain generalization performance results over all 11 classes are depicted in \ref{fig:cross_domain}. Our \textit{VALERIE22} dataset performs best on three datasets (BDD100K, Cityscapes, IDD) and just marginally worse than the \textit{SynPeDS} trained model on A2D2. Compared to the mainly North-American based \textit{Mapillary Vistas} dataset our dataset shows a significant domain shift. Although, still the cross-domain evaluation of \textit{VALERIE22} is significantly better than \textit{Synscapes} and close to \textit{GTAV}.

Most notably our dataset outperforms the \textit{SynPeDS} dataset on the \textit{Cityscapes} dataset. This comes as a surprise as the \textit{SynPeDS} dataset was created to synthetically resemble the \textit{Cityscapes} dataset.


\subsubsection{Number of Assets}

We conducted experiments to understand the influence of diversity of the training data.
Therefore, cross-domain performance is evaluated by comparing the number of unique training assets and the resulting cross-domain segmentation performance.

While comparing automotive real-world and synthetic images it becomes obvious that most images and scenes in real-world images are unique, whereas in synthetic images the scenes are often composed of repetitive content, i.e., a limited amount of unique assets, which are continuously differently arranged. In synthetic datasets the 3D assets, i.e., the 3D meshes and textures of objects in a scene, are expensive to create at a high fidelity and should therefore be used as much as possible. Training a pedestrian detector on a dataset consists of too few unique person assets will lead to a strongly biased detector which is able to detect solely the few trained person assets, but will fail to generalize on other persons.
Overfitting will therefore occur if the training data is of low diversity and the model will fail to generalize, but it is non-obvious on how much diversity is actually needed to generalize well.

To understand the required diversity we investigated the semantic segmentation performance on the \emph{person} class of a \texttt{DeeplabV3+} model trained with different subsets of the \textit{VALERIE22} and the \textit{SynPeDs} datasets.
The subsets, i.e., sequences, of our dataset are described in the Appendix whereas the subsets of the \textit{SynPeDS} dataset, i.e., tranches, are described in \cite{StaunerT2022}.
To track the number of unique person assets per subset in our dataset we just have to count the occurrences of unique asset IDs in the scene metadata files of a sequence.

\begin{figure}[!ht]
  \centering
  \includegraphics[width=1.0\columnwidth]{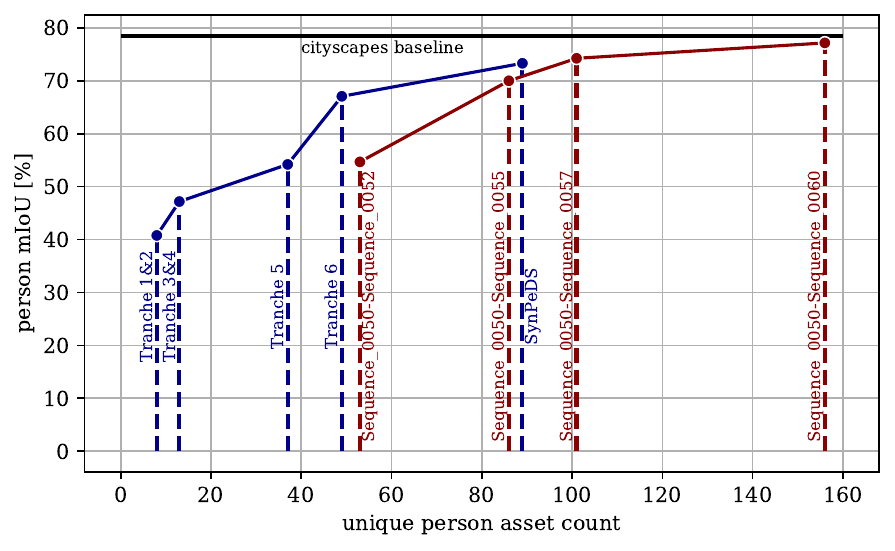}
  \caption{Unique person assets per \textit{SynPeDS} (blue) tranche or \textit{VALERIE22} (red) sequence and person class generalization performance on the \textit{Cityscapes} dataset.}
  \label{fig:assets_miou}
\end{figure}

Each subset of both datasets represents a stage in the process of its development and therefore these dataset subsets consist of an increasing number of pedestrian assets the further the development progressed. The trained models are cross-validated on the \textit{Cityscapes} validation dataset to investigate the cross-domain generalization performance. Figure~\ref{fig:assets_miou} shows the resulting number of unique person assets in the dataset subsets compared to the cross-domain person class performance measured as mIoU on the \textit{Cityscapes} dataset.

The \textit{VALERIE22} subset for higher unique person counts clearly outperforms the \textit{SynPeDS} subset in the cross-domain performance. While a low number of unique assets will lead to overfitting on these assets, a higher number clearly benefits the generalization capabilities of the model. Both, the \textit{VALERIE22} trained models and the \textit{SynPeDS} trained model benefit from an increasing number of person assets on the cross-domain performance. The model trained on our full \textit{VALERIE22} dataset is just $<\%1$ worse in performance than the baseline \textit{Cityscapes} trained model. The results clearly indicate the more diverse a dataset with regard to person assets, the better the generalization capabilities of a segmentation model on this class.

\subsubsection{Number of Training Images}

\begin{figure}[!ht]
  \centering
  \includegraphics[width=1.0\columnwidth]{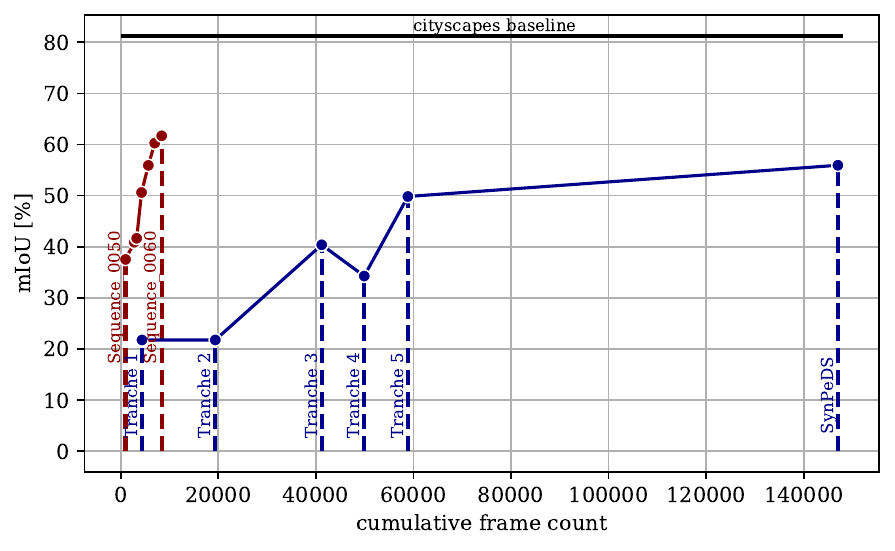}
  \caption{Number of training frames per \textit{SynPeDS} (blue) tranche or \textit{VALERIE22} (red) sequence and overall generalization performance on the \textit{Cityscapes} dataset.}
  \label{fig:framecount_miou}
\end{figure}

Training with a diversified dataset shows significant improvement on the cross-domain performance. This might also raise the question on the performance difference if we have a huge number of training images with lower asset diversity compared to a smaller count of images but with a higher number of assets.
A very low number of images should obviously lead to overfitting, but training with a huge dataset with only marginal differences between images could lead to overfitting as well. From our previous experiment we found that the person asset diversity in the overall \textit{VALERIE22} dataset is higher compared to the \textit{SynPeDS} dataset and this leads to a better segmentation performance. However, the number of training images is vastly different between these datasets. To understand the influence of the number of training images we compared the cross-domain performance on all 11 classes on the \textit{Cityscapes} dataset again trained on subsets of the \textit{VALERIE22} and \textit{SynPeDS} datasets. Figure~\ref{fig:framecount_miou} shows the generalization results with the respective cumulative frame counts that were used to train each segmentation model.

While no model reaches the baseline performance of $82.34\%$, the cross-domain performance with Sequences of our \textit{VALERIE22} dataset reach higher mIoU performance values with far fewer image frames than the \textit{SynPeDS} dataset. As previously shown, the diversity in the \textit{VALERIE22} dataset continuously improved, which is evident by the increasing cross-domain performance, whereas the performance of the \textit{SynPeDS} model even deceased for tranche 4. In tranche 4 a significant pedestrian object distribution bias was introduced into the dataset as was found in \cite{Grau2022}. In \cite{Grau2022} we additionally showed how to utilize the exact positioning metadata of the person assets in the images to identify the pedestrian distributions and understand if data biases were introduced.
Overall, it is clearly visible in this result that only increasing the frame count by reiterating the same assets in the scenes is no viable strategy to increase the cross-domain generalization performance.

\section{Summary}
\label{sec:summary}

This paper describes the \textit{VALERIE22} dataset. The dataset and its underlying scene models are generated completely automated with a parametric scene generation and rendering pipeline.
The results of a cross-evaluation with real and other synthetic datasets demonstrates the performance of this approach. Compared to European datasets VALERIE22 is performing best (or equal) compared with the synthetic \textit{SynPeDS}, \textit{GTAV} and \textit{Synscapes} datasets.

\textit{VALERIE22} comes with a rich set of metadata annotations making it a valuable asset for research on understanding performance and domain aspects of DNNs.

\section*{Acknowledgement}

The work presented in this paper was partially funded by the BMWK project KI Absicherung.

  {\small
    \bibliographystyle{ieee_fullname}
    \bibliography{egbib}

\begin{thebibliography}{10}\itemsep=-1pt

\bibitem{kiaHOME}
KI Absicherung.
\newblock Project home page, https://www.ki-absicherung-projekt.de/en.
\newblock Accessed: 2023-07-20.

\bibitem{Burger1995}
Wilhelm Burger and Matthew~J. Barth.
\newblock {\em Virtual Reality for Enhanced Computer Vision}, pages 247--257.
\newblock Springer US, Boston, MA, 1995.

\bibitem{Carlson_2018_ECCV_Workshops}
Alexandra Carlson, Katherine~A. Skinner, Ram Vasudevan, and Matthew
  Johnson-Roberson.
\newblock Modeling camera effects to improve visual learning from synthetic
  data.
\newblock In {\em Proceedings of the European Conference on Computer Vision
  (ECCV) Workshops}, September 2018.

\bibitem{carlson2019sensor}
Alexandra Carlson, Katherine~A Skinner, Ram Vasudevan, and Matthew
  Johnson-Roberson.
\newblock Sensor transfer: Learning optimal sensor effect image augmentation
  for sim-to-real domain adaptation.
\newblock {\em IEEE Robotics and Automation Letters}, 4(3):2431--2438, 2019.

\bibitem{chen2017deeplab}
Liang-Chieh Chen, George Papandreou, Iasonas Kokkinos, Kevin Murphy, and Alan~L
  Yuille.
\newblock Deeplab: Semantic image segmentation with deep convolutional nets,
  atrous convolution, and fully connected crfs.
\newblock {\em IEEE transactions on pattern analysis and machine intelligence},
  40(4):834--848, 2017.

\bibitem{cordts2016cityscapes}
Marius Cordts, Mohamed Omran, Sebastian Ramos, et~al.
\newblock {The Cityscapes Dataset for Semantic Urban Scene Understanding}.
\newblock In {\em Proceedings of the IEEE/CVF Conference on Computer Vision and
  Pattern Recognition (CVPR)}, pages 3213--3223, Las Vegas, NV, USA, June 2016.

\bibitem{damm2018exploiting}
Werner Damm and Roland Galbas.
\newblock Exploiting learning and scenario-based specification languages for
  the verification and validation of highly automated driving.
\newblock In {\em 2018 IEEE/ACM 1st International Workshop on Software
  Engineering for AI in Autonomous Systems (SEFAIAS)}, pages 39--46. IEEE,
  2018.

\bibitem{devaranjan2020metasim2}
Jeevan Devaranjan, Amlan Kar, and Sanja Fidler.
\newblock Meta-sim2: Learning to generate synthetic datasets.
\newblock In {\em ECCV}, virtual conference, 2020.

\bibitem{DBLP:journals/corr/abs-1711-03938}
Alexey Dosovitskiy, Germ{\'{a}}n Ros, Felipe Codevilla, Antonio L{\'{o}}pez,
  and Vladlen Koltun.
\newblock {CARLA:} an open urban driving simulator.
\newblock {\em CoRR}, abs/1711.03938, 2017.

\bibitem{Unreal4}
Epic Games.
\newblock Unreal engine 4.
\newblock \url{https://www.unrealengine.com}.

\bibitem{aev2019}
Jakob Geyer, Yohannes Kassahun, Mentar Mahmudi, Xavier Ricou, Rupesh Durgesh,
  Andrew~S. Chung, Lorenz Hauswald, Viet~Hoang Pham, Maximilian Mühlegg,
  Sebastian Dorn, Tiffany Fernandez, Martin Jänicke, Sudesh Mirashi,
  Chiragkumar Savani, Martin Sturm, Oleksandr Vorobiov, Martin Oelker,
  Sebastian Garreis, and Peter Schuberth.
\newblock {A2D2: AEV Autonomous Driving Dataset}.
\newblock \url{http://www.a2d2.audi}, 2019.

\bibitem{Grau2022}
Oliver Grau, Korbinian Hagn, and Qutub Syed~Sha.
\newblock {\em A Variational Deep Synthesis Approach for Perception
  Validation}, pages 359--381.
\newblock Springer International Publishing, Cham, 2022.

\bibitem{HagnK2021}
Korbinian Hagn and Oliver Grau.
\newblock Improved sensor model for realistic synthetic data generation.
\newblock In {\em Computer Science in Cars Symposium}, CSCS '21, New York, NY,
  USA, 2021. Association for Computing Machinery.

\bibitem{HagnK2022}
Korbinian Hagn and Oliver Grau.
\newblock Increasing pedestrian detection performance through weighting of
  detection impairing factors.
\newblock In {\em Proceedings of the 6th ACM Computer Science in Cars
  Symposium}, CSCS '22, New York, NY, USA, 2022. Association for Computing
  Machinery.

\bibitem{HagnK2022ECCV}
Korbinian Hagn and Oliver Grau.
\newblock Validation of pedestrian detectors by classification of visual
  detection impairing factors.
\newblock In Leonid Karlinsky, Tomer Michaeli, and Ko Nishino, editors, {\em
  Computer Vision -- ECCV 2022 Workshops}, pages 476--491, Cham, 2023. Springer
  Nature Switzerland.

\bibitem{junietz2018evaluation}
Philipp Junietz, Walther Wachenfeld, Kamil Klonecki, and Hermann Winner.
\newblock Evaluation of different approaches to address safety validation of
  automated driving.
\newblock In {\em 2018 21st International Conference on Intelligent
  Transportation Systems (ITSC)}, pages 491--496. IEEE, 2018.

\bibitem{Kalra16}
Nidhi Kalra and Susan~M. Paddock.
\newblock Driving to safety: How many miles of driving would it take to
  demonstrate autonomous vehicle reliability?, 2016.
\newblock Santa Monica, CA; RAND Corporation.

\bibitem{Liu2020NeuralNG}
Zhenyi Liu, Trisha Lian, J. Farrell, and B. Wandell.
\newblock Neural network generalization: The impact of camera parameters.
\newblock {\em IEEE Access}, 8:10443--10454, 2020.

\bibitem{8954694}
Z. {Liu}, T. {Lian}, J. {Farrell}, and B.~A. {Wandell}.
\newblock Neural network generalization: The impact of camera parameters.
\newblock {\em IEEE Access}, 8:10443--10454, 2020.

\bibitem{Menzel18}
T. {Menzel}, G. {Bagschik}, and M. {Maurer}.
\newblock Scenarios for development, test and validation of automated vehicles.
\newblock In {\em 2018 IEEE Intelligent Vehicles Symposium (IV)}, pages
  1821--1827, June 2018.

\bibitem{MVD2017}
Gerhard Neuhold, Tobias Ollmann, Samuel Rota~Bul\`o, and Peter Kontschieder.
\newblock The mapillary vistas dataset for semantic understanding of street
  scenes.
\newblock In {\em International Conference on Computer Vision (ICCV)}, 2017.

\bibitem{richter2016playing}
Stephan~R. Richter, Vibhav Vineet, Stefan Roth, and Vladlen Koltun.
\newblock Playing for data: Ground truth from computer games.
\newblock In {\em Computer Vision -- ECCV 2016}, pages 102--118, Cham, 2016.
  Springer International Publishing.

\bibitem{shelhamer2016fully}
Evan Shelhamer, Jonathan Long, and Trevor Darrell.
\newblock Fully convolutional networks for semantic segmentation, 2016.

\bibitem{StaunerT2022}
Thomas Stauner, Frederik Blank, Michael F\"{u}rst, Johannes G\"{u}nther,
  Korbinian Hagn, Philipp Heidenreich, Markus Huber, Bastian Knerr, Thomas
  Schulik, and Karl-Ferdinand Lei\ss{}.
\newblock Synpeds: A synthetic dataset for pedestrian detection in urban
  traffic scenes.
\newblock In {\em Proceedings of the 6th ACM Computer Science in Cars
  Symposium}, CSCS '22, New York, NY, USA, 2022. Association for Computing
  Machinery.

\bibitem{SyedShar2020}
Qutub Syed~Sha, Oliver Grau, and Korbinian Hagn.
\newblock Dnn analysis through synthetic data variation.
\newblock In {\em Computer Science in Cars Symposium}, CSCS '20, New York, NY,
  USA, 2020. Association for Computing Machinery.

\bibitem{varma2018idd}
Girish Varma, Anbumani Subramanian, Anoop Namboodiri, Manmohan Chandraker, and
  C~V Jawahar.
\newblock Idd: A dataset for exploring problems of autonomous navigation in
  unconstrained environments, 2018.

\bibitem{wen2020scenario}
Mingyun Wen, Jisun Park, and Kyungeun Cho.
\newblock A scenario generation pipeline for autonomous vehicle simulators.
\newblock {\em Human-centric Computing and Information Sciences}, 10:1--15,
  2020.

\bibitem{wrenninge2018synscapes}
Magnus Wrenninge and Jonas Unger.
\newblock Synscapes: A photorealistic synthetic dataset for street scene
  parsing, 2018.

\bibitem{wymann2000torcs}
Bernhard Wymann, Eric Espi{\'e}, Christophe Guionneau, Christos Dimitrakakis,
  R{\'e}mi Coulom, and Andrew Sumner.
\newblock Torcs, the open racing car simulator.
\newblock {\em Software available at http://torcs. sourceforge. net}, 4(6):2,
  2000.

\bibitem{Yu2019}
Fisher Yu, Haofeng Chen, Xin Wang, Wenqi Xian, Yingying Chen, Fangchen Liu,
  Vashisht Madhavan, and Trevor Darrell.
\newblock {BDD100K: A Diverse Driving Dataset for Heterogeneous Multitask
  Learning}.
\newblock In {\em Proc. of CVPR}, pages 1--14, Seattle, WA, USA, June 2020.

\bibitem{yurtsever2020survey}
Ekim Yurtsever, Jacob Lambert, Alexander Carballo, and Kazuya Takeda.
\newblock A survey of autonomous driving: Common practices and emerging
  technologies.
\newblock {\em IEEE Access}, 8:58443--58469, 2020.

\end{thebibliography}
  }

\section*{Appendix}
\label{sec:appendix}

\lstset{
  basicstyle=\ttfamily,
  columns=fullflexible,
  frame=single,
  breaklines=true,
  postbreak=\mbox{\textcolor{red}{$\hookrightarrow$}\space},
}

The VALERIE22 dataset is structured in sequence groups. Each group has certain characteristics, like the amount of different assets used and scene complexity and composition. Each sequence contains images tone-mapped with the default Blender tone-mapper (in the "png" folder) and the same images distorted with our sensor simulation as described in section~\ref{sec:sensor_simulation} (in the "png\_distorted" folder).
The following gives  a brief description of the main sequence groups.

\FloatBarrier
\subsection*{Sequence 50}
\label{seq:sequence_50}
\begin{figure}[hbt!]
  \centering
  \includegraphics[width=0.95\linewidth]{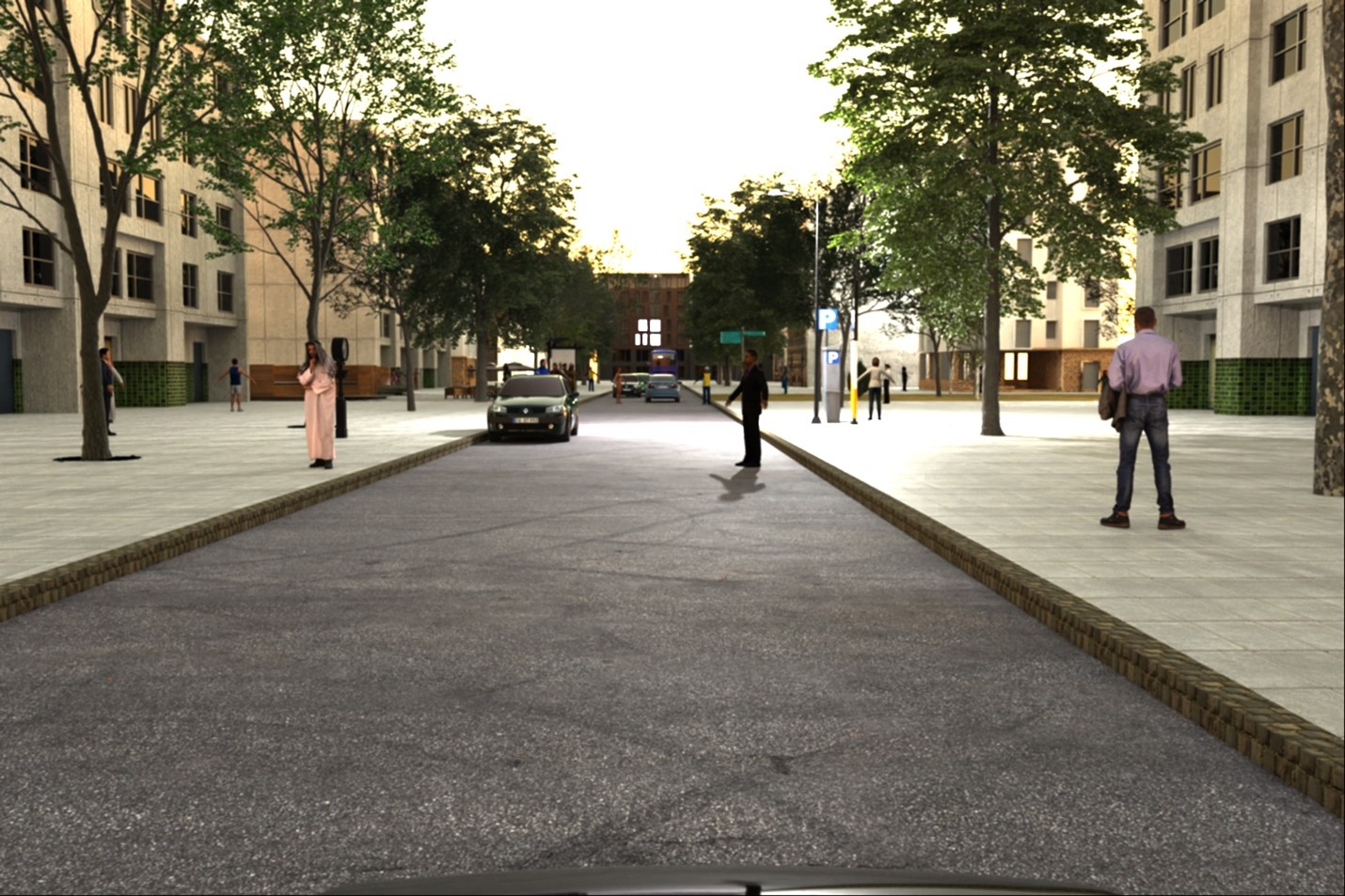}
  \caption{Example frame sequence 50.}
  \label{fig:seq50}
\end{figure}

Based on a simple grammar (Valcity-1) with a fixed street and one crossing. It contains some night scenes with artificial light sources.

\FloatBarrier
\subsection*{Sequence 52}
\label{seq:sequence_52}
\begin{figure}[hbt!]
  \centering
  \includegraphics[width=0.95\linewidth]{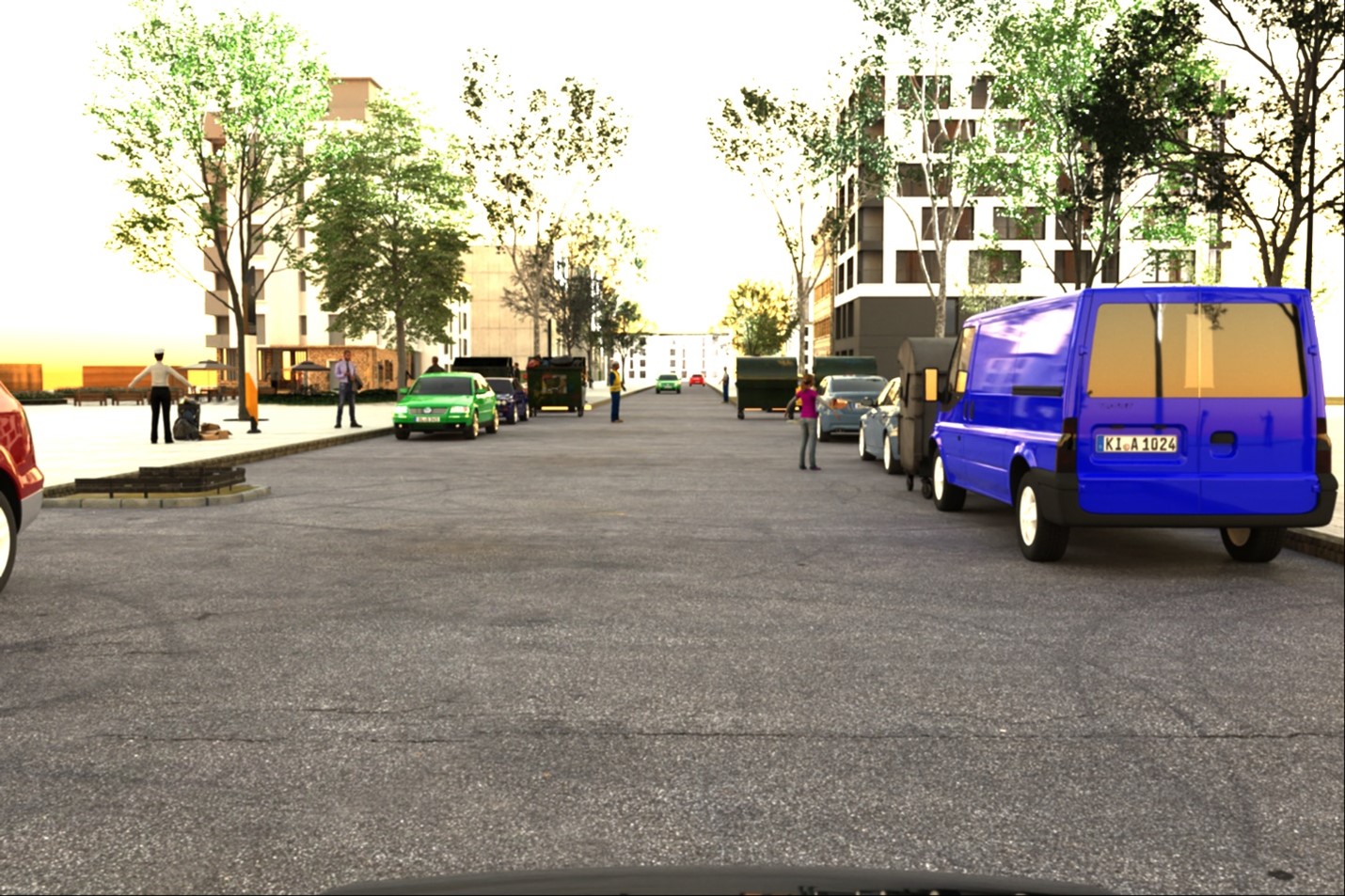}
  \caption{Example frame sequence 52.}
  \label{fig:seq52}
\end{figure}

Similar to seq 50, new assets. Street base with two crossings, includes night lit scenes.
The time of the day is varied  5:30am to 9pm, with some artificial light in the late hours.

\FloatBarrier
\subsection*{Sequence 54}
\label{seq:sequence_54}
\begin{figure}[hbt!]
  \centering
  \includegraphics[width=0.95\linewidth]{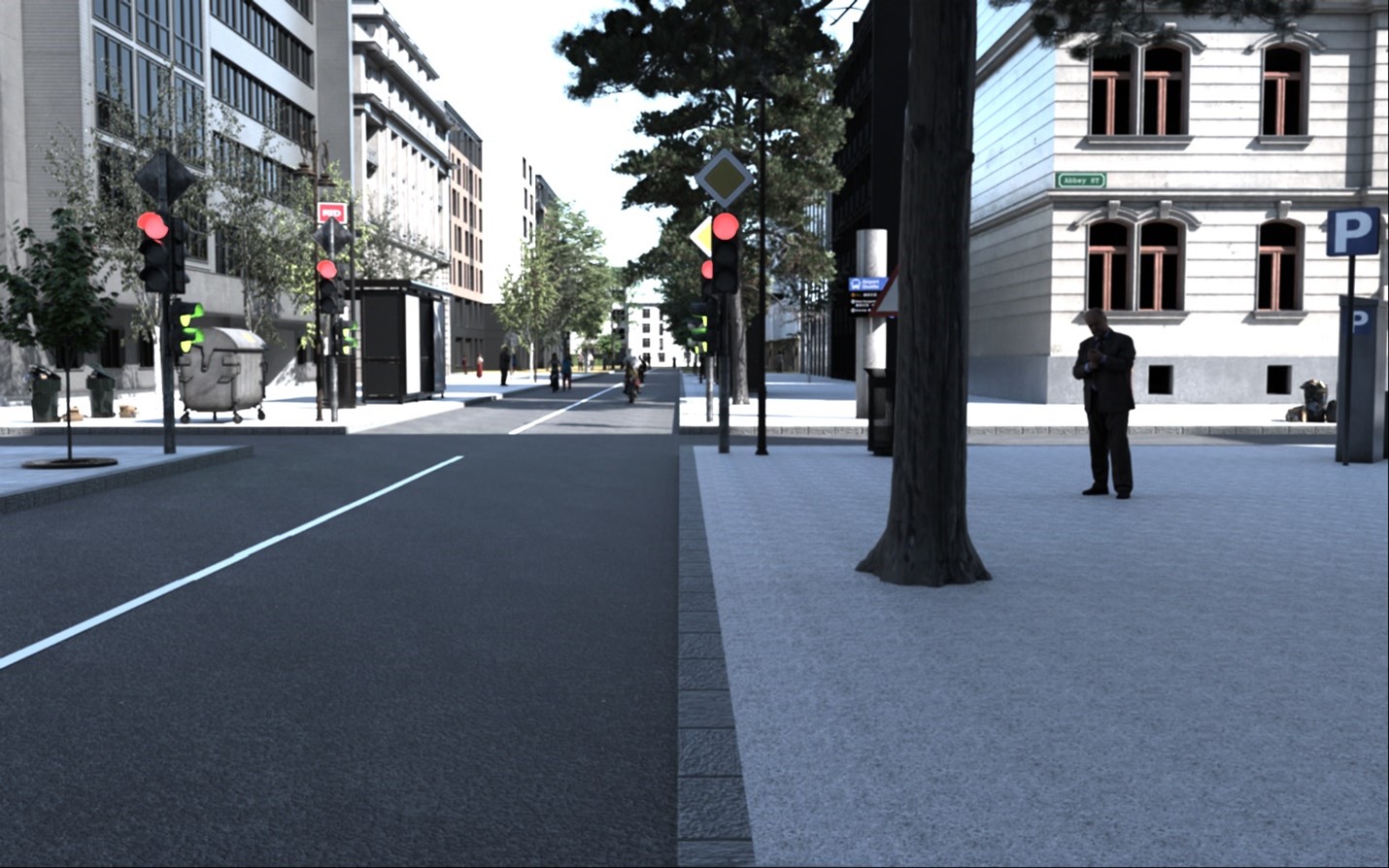}
  \caption{Example frame sequence 54.}
  \label{fig:seq54}
\end{figure}

Based on Valeriecity scene generation engine. Has a generic 2 crossing street with variation of the main street width.
Includes few traffic signs, low-mid person density. Time: 10:05am.

This version of the scene generator uses an automatic layout of the street (autolane feature). Depending on the width it includes park lanes and separate lanes in each direction.

Excerpt from the scene generator configuration file:

\begin{lstlisting}
 "facade_dist": 4,
 "static_stuff_distance": 1,
 "density_road_persons": 0.005,
 "stdvar_road_persons": 0.001,
 "density_side_persons": 0.01,
 "stdvar_side_persons": 0.005,
 "density_lane_cars": 0.005,
 "stdvar_lane_cars": 0.002,
 "density_parking_cars": 0.003,
 "stdvar_parking_cars": 0.01,
 "parking_separator_spacing": 25,
 "tree_density": 0.01,
 "tree_density_sidewalk": 0.01,
 "parking_separator_var": 2.0
\end{lstlisting}

\FloatBarrier
\subsection*{Sequence 57}
\label{seq:sequence_57}
\begin{figure}[hbt!]
  \centering
  \includegraphics[width=0.95\linewidth]{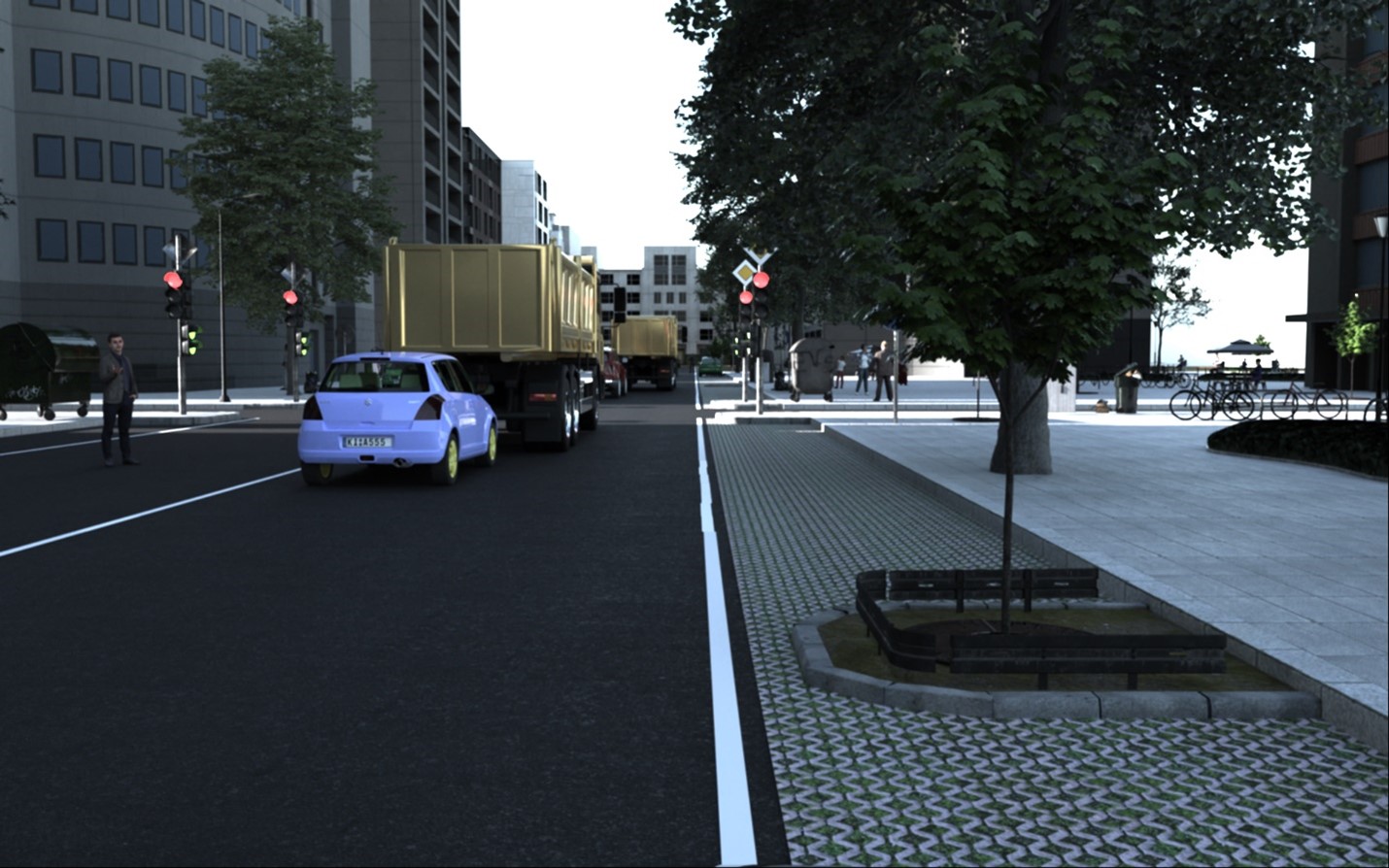}
  \caption{Example frame sequence 57.}
  \label{fig:seq57}
\end{figure}

Based on previous settings, these scenes contain generic 2 crossing street. The scene generator was generation variation of 7 sidewalk distances (1..15m), 150 street width variation (3..24m) and time-of-the-day 6:00-20:00.

\begin{lstlisting}
"density_road_persons": 0.005,
 "stdvar_road_persons": 0.001,
 "density_side_persons": 0.01,
 "stdvar_side_persons": 0.005,
 "density_lane_cars": 0.005,
 "stdvar_lane_cars": 0.002,
 "density_parking_cars": 0.003,
 "stdvar_parking_cars": 0.01,
 "tree_density": 0.01,
 "tree_density_sidewalk": 0.01,
\end{lstlisting}

\FloatBarrier
\subsection*{Sequence 58}
\label{seq:sequence_58}
\begin{figure}[hbt!]
  \centering
  \includegraphics[width=0.95\linewidth]{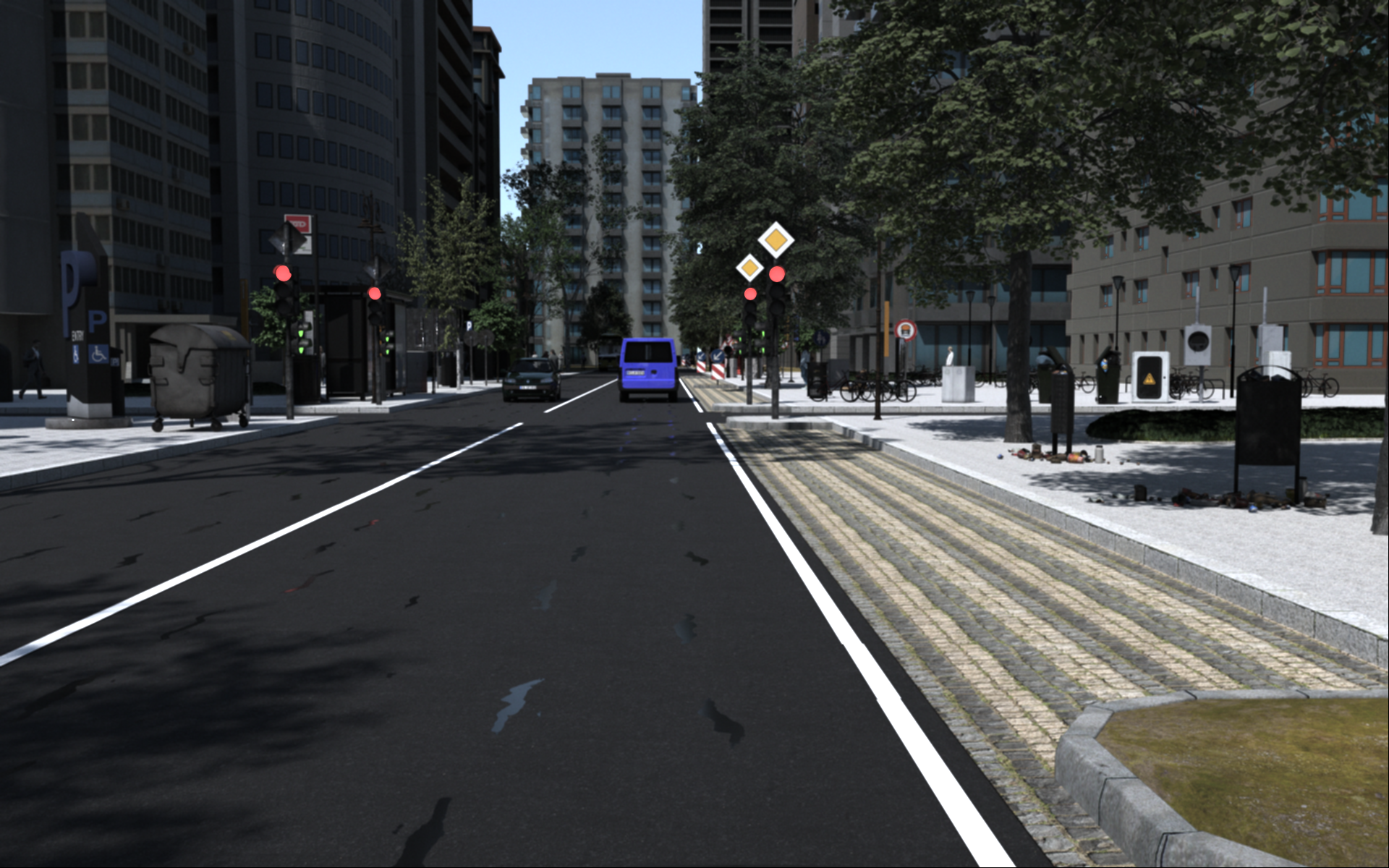}
  \caption{Example frame sequence 58.}
  \label{fig:seq58}
\end{figure}

Like sequence 57, random sampled time 6:30-20:00, 1 time sampled.

\FloatBarrier
\subsection*{Sequence 59}
\label{seq:sequence_59}
\begin{figure}[hbt!]
  \centering
  \includegraphics[width=0.95\linewidth]{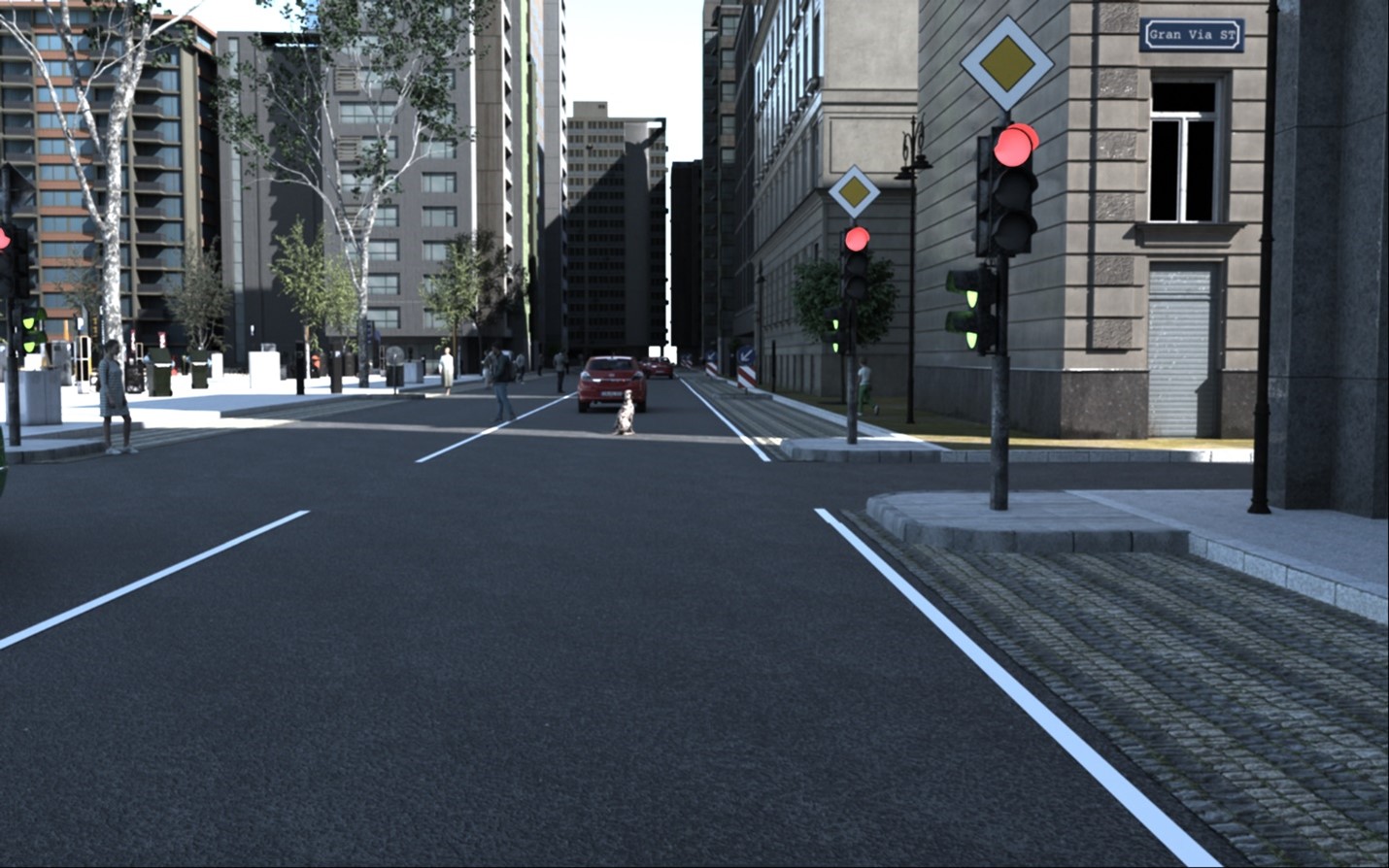}
  \caption{Example frame sequence 59.}
  \label{fig:seq59}
\end{figure}

Like sequence 57.  5 sidewalk distances (1..15m), 150 street width variations [5..25m], time 10:30, two ego-vehicle position (North and South looking).

\FloatBarrier
\subsection*{Sequence 60}
\label{seq:sequence_60}
\begin{figure}[hbt!]
  \centering
  \includegraphics[width=0.95\linewidth]{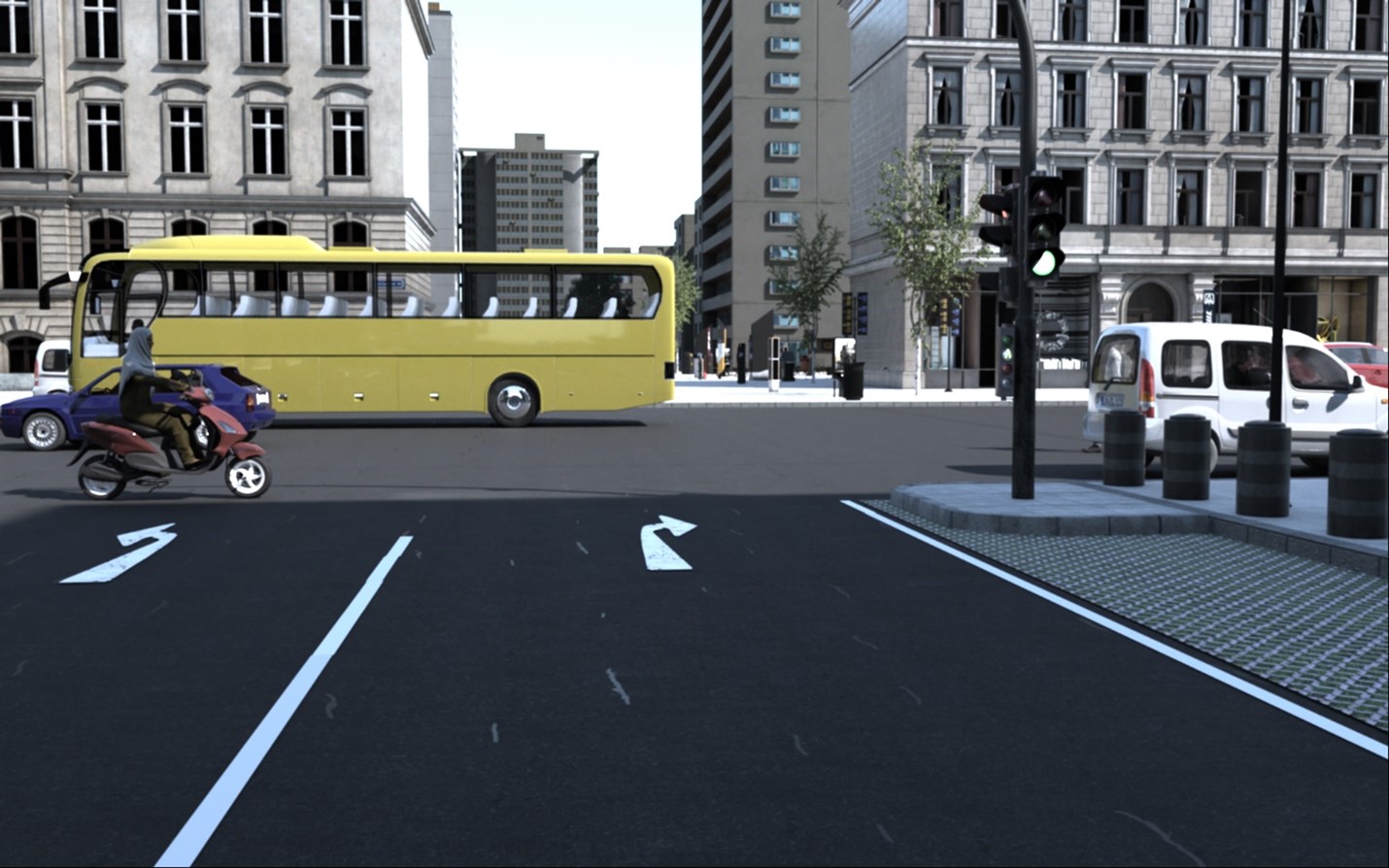}
  \caption{Example frame sequence 60.}
  \label{fig:seq60}
\end{figure}

A generic crossing and T-junction, 4 sidewalk distances (2..15m), 100 width variations [12..37m], with middle ground on wider street widths, time 10:30, two ego-vehicle position (North and South looking), not all possible frames rendered, ego-car rotation random sampled in [155..210], [-15.0, 15.0].

\FloatBarrier
\subsection*{Sequence 62}
\label{seq:sequence_62}
\begin{figure}[hbt!]
  \centering
  \includegraphics[width=0.95\linewidth]{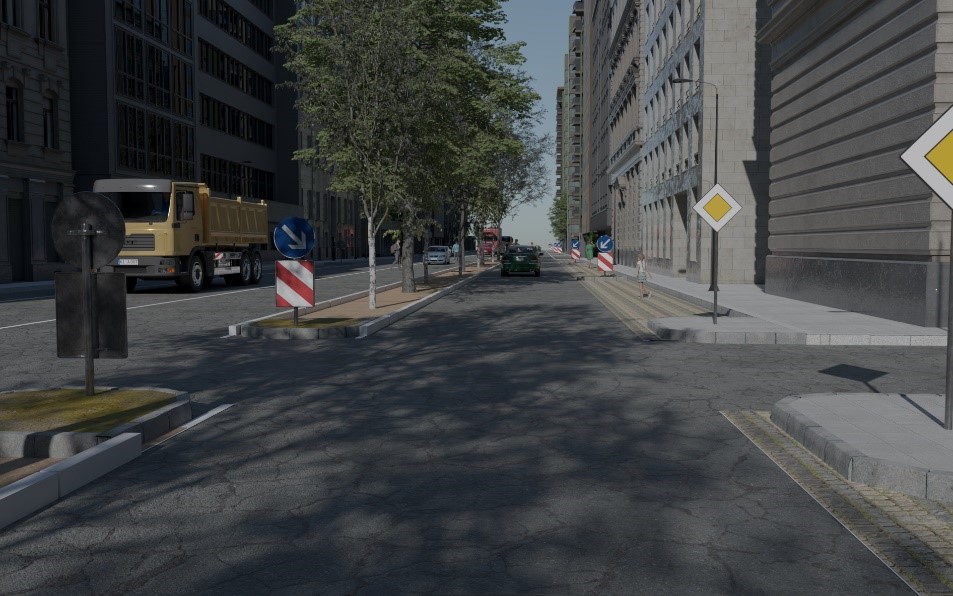}
  \caption{Example frame sequence 62.}
  \label{fig:seq62}
\end{figure}

Based on seq57 scenegen parameter. Variation in time 7:00 + 10:12 and North-offset [80..110] 4 samples, not all variations rendered.

\clearpage

\FloatBarrier
\section*{Ground Truth and Metadata}
\label{sec:ground_truth}

This section describes the ground truth annotations available in our \textit{VALERIE22} dataset.

\FloatBarrier
\subsection*{2d-bounding-box\_json}
\label{sec:2d_bounding_box}

The 2d-bounding-box\_json directory contains the 2d bounding box annotations for the person objects. Every person object entry has a \{Entity-ID\}, i.e., an object identifier that matches with the "Entities" ID in the respective \hyperref[sec:general_globally]{\textbf{general-globally-per-frame-analysis\_json}} file. The bounding boxes are defined with their center coordinates, width and height. There are two bounding box entries per person, one defines the visible person object and the other the unoccluded person object.
For every entry there are additional metadata entries, i.e., performance limiting factors, such as the occlusion rate, distance to the camera in meters etc.

\begin{lstlisting}
{
  "{Entity-ID}": {
    "class_id": "person",
    "bb": {
        "c_x": 438.5,
        "c_y": 504.0,
        "w": 13,
        "h": 46
    },
    "bb_vis": {
        "c_x": 433.0,
        "c_y": 498.5,
        "w": 2,
        "h": 3
    },
    "occlusion": 0.9809523809523809,
    "distance": 69.94487874538726,
    "v_x": 0,
    "v_y": 0,
    "truncated": false,
    "total_pixels_object": "315",
    "total_visible_pixels_object": "6",
    "occluder_ids": "342",
    "occluder_pixels": "309",
    "contrast_rgb_full": 36.8348,
    "contrast_edge": 36.8348,
    "contrast_rgb": 17.81663,
    "luminance": 0.14153883892191005,
    "perceived_lightness": 43.74599160710958
  }
}
\end{lstlisting}

\FloatBarrier
\subsection*{3d-bounding-box\_json}
\label{sec:3d_bounding_box}

The 3d-bounding-box\_json directory contains the 3d bounding box annotations for every object. Every entry has a \{Entity-ID\} that matches with the "Entities" ID in the respective \hyperref[sec:general_globally]{\textbf{general-globally-per-frame-analysis\_json}} file. The bounding boxes are defined with their center coordinates, x, y and z dimensions in meters.

\begin{lstlisting}
{
  "{Entity-ID}": {
    "obj_id": "60c39f28-bf2a-11ec-bcf0-0242ac110003",
    "class_id": "person",
    "center": [
        110.20051956176758,
        78.18697738647461,
        0.877331018447876
    ],
    "size": [
        0.8750991821289062,
        1.0738754272460938,
        1.754662036895752
    ]
  }
}
\end{lstlisting}

\clearpage

\FloatBarrier
\subsection*{class-id\_png}
\label{sec:class_id}

The class-id\_png files contain the \hyperref[sec:semantic_group_segmentation]{\textbf{semantic-group-segmentation\_png}} files mapped to the following 11 classes in grayscale.

\begin{table}[th]
  \caption{Ground truth class mapping of semantic group segmentation to trainIds and color.}
  \label{tab:semantic_group_segmentation}
  \centering
  \begin{tabular}{l|c|c|l}
    \hline
    class                & id & trainId & color         \\
    \hline
    \hline
    unlabeled            & 0  & 255     & [  0,  0,  0] \\
    ego vehicle          & 1  & 255     & [  0,  0,  0] \\
    rectification border & 2  & 255     & [  0,  0,  0] \\
    out of roi           & 3  & 255     & [  0,  0,  0] \\
    static               & 4  & 255     & [  0,  0,  0] \\
    dynamic              & 5  & 255     & [111, 74,  0] \\
    ground               & 6  & 255     & [ 81,  0, 81] \\
    road                 & 7  & 0       & [128, 64,128] \\
    sidewalk             & 8  & 1       & [244, 35,232] \\
    parking              & 9  & 255     & [250,170,160] \\
    rail track           & 10 & 255     & [230,150,140] \\
    building             & 11 & 2       & [ 70, 70, 70] \\
    wall                 & 12 & 255     & [102,102,156] \\
    fence                & 13 & 255     & [190,153,153] \\
    guard rail           & 14 & 255     & [180,165,180] \\
    bridge               & 15 & 255     & [150,100,100] \\
    tunnel               & 16 & 255     & [150,120, 90] \\
    pole                 & 17 & 3       & [153,153,153] \\
    polegroup            & 18 & 3       & [153,153,153] \\
    traffic light        & 19 & 4       & [250,170, 30] \\
    traffic sign         & 20 & 5       & [220,220,  0] \\
    vegetation           & 21 & 6       & [107,142, 35] \\
    terrain              & 22 & 6       & [152,251,152] \\
    sky                  & 23 & 7       & [ 70,130,180] \\
    person               & 24 & 8       & [220, 20, 60] \\
    rider                & 25 & 8       & [255,  0,  0] \\
    motorcycle           & 32 & 255     & [  0,  0,230] \\
    bicycle              & 33 & 255     & [119, 11, 32] \\
    car                  & 26 & 9       & [  0,  0,142] \\
    truck                & 27 & 10      & [  0,  0, 70] \\
    bus                  & 28 & 10      & [  0, 60,100] \\
    caravan              & 29 & 255     & [  0,  0, 90] \\
    trailer              & 30 & 255     & [  0,  0,110] \\
    train                & 31 & 10      & [  0, 80,100] \\
    \hline
  \end{tabular}
\end{table}

\FloatBarrier
\subsection*{semantic-group-segmentation\_png}
\label{sec:semantic_group_segmentation}

The semantic-group-segmentation\_png files contain the segmentation [RGB] files with class to color mapping as defined in Table~\ref{tab:semantic_group_segmentation}.

\FloatBarrier
\subsection*{semantic-instance-segmentation\_png}
\label{sec:semantic_instance_segmentation}

The semantic-instance-segmentation\_png files contain the instance segmentation labels for each frame. The instances are encoded as $\{Entity-ID\} = R * 2^{16} + G * 2^{8} + B$.

For every instance segmentation file there is a subdirectory with the same name. This subdirectory contains \{Entity-ID\}.png files. The files contain the unoccluded entity (person) instance of the corresponding semantic-instance-segmentation file.

\FloatBarrier
\subsection*{general-globally-per-frame-analysis\_json}
\label{sec:general_globally}

The general globally per frame analysis file defines the Environment, Camera and Entities, i.e., Objects, per frame.

\begin{lstlisting}
{
  "__version_entry__": [
    {
      "__Version__": "1.9",
      "__Tool__": "valerie/VRender (Intel)",
      "__Time__": "Wed, 15 Jun 2022 10:24:30 +0000"
    }
  ],
  "meta": {
    "annotator": "Intel"
  },
  "Environment": {
    "__SkyModel__": {
      "time": [
        2020,
        4,
        25,
        10.5
      ],
      "Latitude": 48.135101318359375,
      "Longitude": 11.581999778747559,
      "direction": [
        0.6302990317344666,
        -0.39718276262283325,
        0.6670599579811096
      ],
      "weather": "clear"
    }
  },
  "Camera": {
    "name": "Camera#9247fe66-5ff0-4a38-9564-3b4729f271cf",
    "pos": [
      91.36333465576172,
      173.92953491210938,
      1.429999828338623
    ],
    "rot": [
      86.50000866808853,
      -0.0,
      6.760135334790515
    ],
    "A": [
      -0.11749350279569626,
      0.9911954402923584,
      -0.06104838848114014
    ],
    "Up": [
      -0.0071861925534904,
      0.060623958706855774,
      0.9981347918510437
    ],
    "focallen": 0.03247285842895508,
    "Size": [
      0.036,
      0.024
    ],
    "clipstart": 0.10000000149011612,
    "clipend": 538.8877563476562
  },
  "Entities": {
    "{Entity-ID}": {
      "obj_id": "b0027ec8-d002-4914-988c-23a284cddf44",
      "class_id": "void",
      "obj_name": "void#b0027ec8-d002-4914-988c-23a284cddf44",
      "pos": [
        91.48575592041016,
        172.89675903320312,
        0.0
      ],
      "rot": [
        0.0,
        0.0,
        186.7601300984631
      ],
      "distance_to_camera": 1.7681946983115833,
      "angle_to_camera": 53.97246217995341,
      "A": [
        -0.11771297454833984,
        0.9930476546287537,
        0.0
      ],
      "prototype": {
        "reference": "catalogue",
        "asset_id": "4999474d-50da-4901-b2b5-cf5bc24172d1"
      }
    },
		..
}
\end{lstlisting}

\subsection*{Scene metadata file}
\label{sec:metadata_appendix}

The scene metadata file defines the sampling parameter used to create the scene and frame.

\begin{lstlisting}
{
    "scene_meta_data": null,
    "variant": {
        "__Node__": true,
        "Name": "node-1",
        "attributes": [
            {
                "SampleF": {
                    "__SampleF__": true,
                    "Name": "scene.sun.hrs",
                    "index": 0,
                    "v": 19.495632085834693,
                    "vmin": 6.5,
                    "vmax": 20.0,
                    "step": 1,
                    "samplemode": 1
                }
            }
        ],
        "uuid": "ee44c412-b5ca-11ec-8d3d-a0369f7557f6",
        "vindex": 0
    }
}
\end{lstlisting}

\clearpage

\section*{Dataset Directory Structure}
\label{sec:directory}

\dirtree{%
  .1 \textit{VALERIE22}.
  .2 \hyperref[seq:sequence_50]{\textbf{intel\_results\_sequence\_0050}}.
  .3 \hyperref[sec:ground_truth]{\textbf{ground-truth}}.
  .4 \hyperref[sec:2d_bounding_box]{\textbf{2d-bounding-box\_json}}.
  .5 car-camera000-0000-\{UUID\}-0000.json.
  .4 \hyperref[sec:3d_bounding_box]{\textbf{3d-bounding-box\_json}}.
  .5 car-camera000-0000-\{UUID\}-0000.json.
  .4 \hyperref[sec:class_id]{\textbf{class-id\_png}}.
  .5 car-camera000-0000-\{UUID\}-0000.png.
  .4 \hyperref[sec:general_globally]{\textbf{general-globally-per-frame-analysis\_json}}.
  .5 car-camera000-0000-\{UUID\}-0000.json.
  .5 car-camera000-0000-\{UUID\}-0000.csv.
  .4 \hyperref[sec:semantic_group_segmentation]{\textbf{semantic-group-segmentation\_png}}.
  .5 car-camera000-0000-\{UUID\}-0000.png.
  .4 \hyperref[sec:semantic_instance_segmentation]{\textbf{semantic-instance-segmentation\_png}}.
  .5 car-camera000-0000-\{UUID\}-0000.png.
  .5 car-camera000-0000-\{UUID\}-0000.
  .6 \{Entity-ID\}.
  .3 \hyperref[sec:metadata_appendix]{\textbf{metadata}}.
  .4 car-camera000-0000-\{UUID\}-0000.json.
  .3 sensor.
  .4 camera.
  .5 left.
  .6 png.
  .7 car-camera000-0000-\{UUID\}-0000.png.
  .6 png\_distorted.
  .7 car-camera000-0000-\{UUID\}-0000.png.
  .2 \hyperref[seq:sequence_52]{\textbf{intel\_results\_sequence\_0052}}.
  .2 \hyperref[seq:sequence_54]{\textbf{intel\_results\_sequence\_0054}}.
  .2 \hyperref[seq:sequence_57]{\textbf{intel\_results\_sequence\_0057}}.
  .2 \hyperref[seq:sequence_58]{\textbf{intel\_results\_sequence\_0058}}.
  .2 \hyperref[seq:sequence_59]{\textbf{intel\_results\_sequence\_0059}}.
  .2 \hyperref[seq:sequence_60]{\textbf{intel\_results\_sequence\_0060}}.
  .2 \hyperref[seq:sequence_62]{\textbf{intel\_results\_sequence\_0062}}.
}

\end{document}